\documentclass[runningheads]{llncs}
\usepackage{graphicx}
\usepackage{amsmath,amssymb} 
\usepackage{multirow}
\usepackage{color}
\usepackage[outercaption]{sidecap}
\usepackage[width=122mm,left=12mm,paperwidth=146mm,height=193mm,top=12mm,paperheight=217mm]{geometry}
\usepackage{makecell}

\begin{document}
\pagestyle{headings}
\mainmatter

\title{Deep Multiple Instance Learning for Zero-shot Image Tagging} 

\titlerunning{Deep Multiple Instance Learning for ZST}

\authorrunning{Shafin Rahman and Salman Khan}

\author{Shafin Rahman$^{\dagger\star}$ and Salman Khan$^{\star\dagger}$}


\institute{$^{\dagger}$Australian National University\\
	$^{\star}$DATA61, CSIRO
}

\maketitle

\begin{abstract}
In-line with the success of deep learning on traditional recognition problem, several end-to-end deep models for zero-shot recognition have been proposed in the literature. These models are successful to predict a single unseen label given an input image, but does not scale to cases where multiple unseen objects are present. In this paper, we model this problem within the framework of Multiple Instance Learning (MIL). To the best of our knowledge, we propose the first end-to-end trainable deep MIL framework for the multi-label zero-shot tagging problem. Due to its novel design, the proposed framework has several interesting features: $(1)$ Unlike previous deep MIL models, it does not use any off-line procedure (e.g., Selective Search or EdgeBoxes) for bag generation. $(2)$ During test time, it can process any number of unseen labels given their semantic embedding vectors. $(3)$ Using only seen labels per image as weak annotation, it can produce a bounding box for each predicted labels. We experiment with the NUS-WIDE dataset and achieve superior performance across conventional, zero-shot and generalized zero-shot tagging tasks.
\end{abstract}

\section{Introduction}

In recent years, numerous single label zero-shot classification methods have been proposed aiming to recognize novel concepts with no training examples \cite{Xian_CVPR_2017,Kodirov_2017_CVPR,Zhang_2017_CVPR,Deutsch_2017_CVPR,Li_2017_CVPR,Morgado_2017_CVPR,Akata_2016_CVPR}. However in real life settings, images often come with multiple objects or concepts that may or may not be observed during training. For example, enormous growth in on-line photo collections require automatic image tagging algorithms that can provide both seen and unseen labels to the images.  Despite the importance and practical nature of this problem, there are very few existing methods with the capability to address the zero-shot image tagging task \cite{Mensink_CVPR_2014,Fu_Transductive_2015,Zhang_2016_CVPR,Li_tagging_2015}. Notably, any object or concept can either be present at the localized region or be inferred from the holistic scene information. Moreover, part of the object or concept can be partially occluded by other objects. Therefore, assigning the multiple tags to an image requires searching for both local and global details present at different scales, orientations, poses and illumination conditions. It is considered as a challenging task even for the traditional case let alone the zero-shot version of the problem. 

The existing attempts on zero-shot image tagging (e.g., \cite{Mensink_CVPR_2014,Fu_Transductive_2015,Zhang_2016_CVPR,Li_tagging_2015}) mostly used deep CNN features of the whole images. But, as objects or concepts are more likely to be present at the localized regions, the extracted features from the whole image cannot represent all possible labels. In this paper, we propose a deep Multiple Instance Learning (MIL) framework that operates on a bag of instances generated from each image.  MIL assumes that each bag contains at least one instance of the true labels defined by the ground-truth. A distinguishing feature of our approach is that the bag of instances not only encodes the global information about a scene, but also have a rich representation for localized object-level features. This is different from the existing literature in image tagging, where MIL based strategies do not consider local image details. For example, \cite{Feng_AAAI_2017} used image-level features at different scale of the network whereas \cite{Zhou_2017_ICCV} used whole image features from different video frames as the bag of instance. These techniques therefore work only for the most prominent objects but often fail for non-salient concepts due to the lack of localized information.

In addition to the use of localized features, we integrate the instance localization within a unified system that can be trained jointly in an end-to-end manner. We note that the previous approaches \cite{Wang_ICCV_2015,Wu_CVPR_2015,Wei_PAMI_2016,TANG_PR_2017} applied off-line procedures like Selective Search \cite{SS_2013}/EdgeBoxes \cite{Edge_Boxes_2014}/BING \cite{cheng_2014_bing} for object proposal or patch generation which served as a bag-of-instance. Patches obtained from any external procedure were used in three ways: (1) to extract features for each patch from a pre-trained deep network to create bag-of-instance \cite{Wang_ICCV_2015,Tang_TIP_2017}, (2) to use a set of patches as a bag and feed the bag as an input to a deep network to perform MIL such that the image features can be fine-tined \cite{Wu_CVPR_2015,Wei_PAMI_2016}, (3) to feed an input image with all patch locations and later perform a ROIPooling to generate bag of features \cite{TANG_PR_2017,FastRCNN_2015_ICCV}. One common issue with these approaches is that none of the methods are end-to-end trainable because of the dependency on the external non-differentiable procedure in either feature extraction or bag of instance generation. Furthermore, the above methods do not address zero-shot tagging problem and cannot relate the visual and semantic concepts. A recent work from Ren et al. \cite{Ren_BMVC_17} proposed a solution for zero-shot tagging in-line with the third approach above, that cannot dynamically adapt the patch locations.

\begin{figure}[t] 
  \centering
   \includegraphics[width=1\textwidth,trim={0cm 0cm 0cm 0cm},clip]{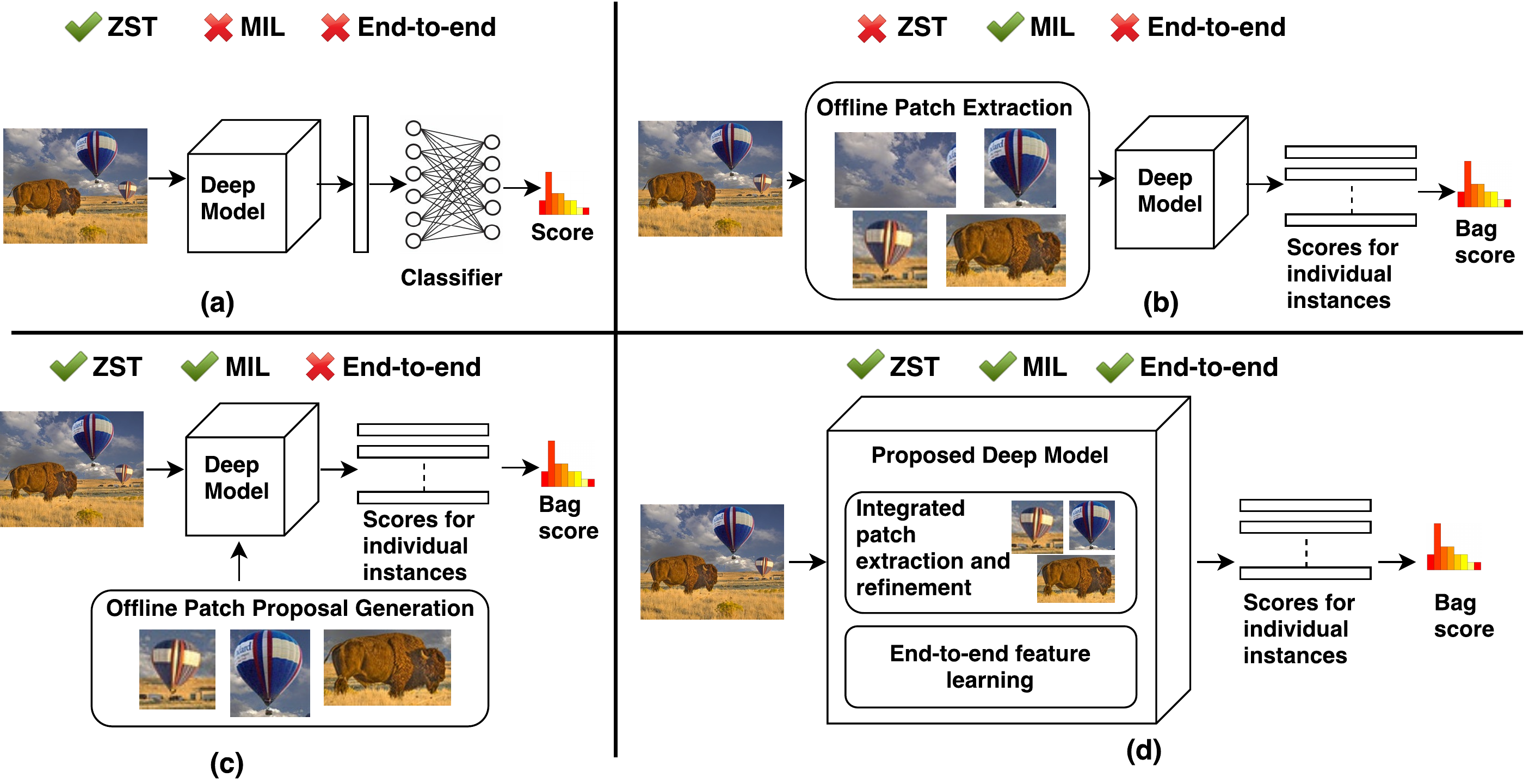}
   \vspace{-0.5em}
   \caption{Overview of different multi-label image annotation architecture. (a) \cite{Zhang_2016_CVPR,Wang_ICCV_2015,Tang_TIP_2017} extract deep features separately, then feed those features to a neural network for classification (b) \cite{Wu_CVPR_2015,Wei_PAMI_2016} use external procedure to extract image patches for bag generation then feed those patches to a deep model to get final bag score for seen classes (c) \cite{TANG_PR_2017,Ren_BMVC_17,FastRCNN_2015_ICCV} process the whole image as well as some patches obtained using external process together where all patches from one image share the same CNN layers to get bag score for seen classes (d) Our proposed MIL model simply takes an image as an input and produces bag score for both seen and unseen classes.}
\label{fig:overview}
\end{figure}

Finally, our proposed network maps local and global information inside the bag to a semantic embedding space so that the correspondences with both seen and unseen classes can be found. 
With respect to zero-shot tagging, \cite{Zhang_2016_CVPR,Ren_BMVC_17} are closest to our work as they handle a visual-semantic mapping to bridge seen and unseen classes. However, \cite{Zhang_2016_CVPR} uses off the shelf features and does not consider local details. To extract local image information, our approach finds localized patches similar to \cite{Ren_BMVC_17}. However, instead of using an external off-line procedure as in \cite{Ren_BMVC_17} for bag generation, our network produces the bag by itself. Moreover, it can simultaneously fine-tune the feature extraction and bag generation process instead of just using pre-trained patch features. In Figure \ref{fig:overview}, we illustrate a block diagram of different kinds of image tagging frameworks. Our framework encompasses all different modules of MIL into only a single integrated network, which results in state-of-the-art performances for conventional and zero-shot image tagging. 



The key contributions of our paper are:
\begin{itemize}
\item We propose the first truly end-to-end deep MIL framework for multi-label image tagging that can work in both conventional and zero-shot settings.

\item The proposed method does not require any off-line procedure to generate bag of instances for MIL, rather incorporates this step in a joint framework.

\item The proposed method is extendable to any number of novel tags from an open vocabulary as it does not use any prior information about the unseen concept.

\item The proposed method can annotate a bounding box for both seen and unseen tags without even using any ground-truth bounding box during training.
\end{itemize}

\section{Related Work}

\subsubsection{Zero-shot learning:} In recent years, we have seen some exciting papers on Zero Shot Learning (ZSL).
The overall goal of those efforts is basically to classify an image to an unseen class for which no training is performed. Investigations are focused on domain adaptation \cite{Deutsch_2017_CVPR}, class attribute association \cite{Demirel_2017_ICCV}, unsupervised semantics \cite{Deutsch_2017_CVPR}, hubness effect \cite{Zhang_2017_CVPR} and generalized setting \cite{Xian_CVPR_2017} of inductive \cite{Kodirov_2017_CVPR} or transductive ZSL learning \cite{Li_2017_CVPR}. The major shortcoming of these approaches is their inability to assign multiple labels to an image, which is a major limitation in real-world settings. In line with the general consideration of only a singe label per image, traditional ZSL methods use recognition datasets which mostly contain only one prominent concept per image. Here, we present an end-to-end deep zero shot tagging method that can assign multiple tags per image.

\subsubsection{End-to-end object detection:} Image tags can corresponds to either whole image or any specific location inside it. To model the relations between tag and its corresponding locations we intend to localize objects in a scene. To do so, we are interested in end-to-end object detection framework. Popular examples of such frameworks are Faster R-CNN \cite{Faster_RCNN_2017}, R-FCN \cite{Dai_RFCN_2016}, SSD \cite{Liu_SSD_2016} and YOLO \cite{Redmon_yolo9000_2016}. The main distinction among these models is the object localization process. R-CNN \cite{Faster_RCNN_2017} and R-FCN \cite{Dai_RFCN_2016} used a Region Proposal Network (RPN) to generate object proposals whereas SSD \cite{Liu_SSD_2016} and YOLO \cite{Redmon_yolo9000_2016} propose bounding box and classify it in a single step. The later group of models usually runs faster but relatively less accurate than the first group. Due to the focus on highly accurate object detection, we built on Faster R-CNN \cite{Faster_RCNN_2017} as the backbone architecture in the current work.

\subsubsection{Image tagging with deep MIL:} Existing image tagging approaches based on MIL mostly use hand crafted or deep CNN features \cite{Tang_TIP_2017,Wang_ICCV_2015}. End-to-end learnable architectures for image tagging are relatively less explored in the literature. Based on bag-of-instance generation, these architectures are of two kinds. The \emph{first} kind generates bags using an offline procedure, external to the deep architecture. For example, \cite{Wu_CVPR_2015,Wei_PAMI_2016} generate object proposal/patches using Selective Search \cite{SS_2013}/EdgeBoxes \cite{Edge_Boxes_2014}/BING \cite{cheng_2014_bing} and feed those into the network separately. In contrast, \cite{FastRCNN_2015_ICCV,TANG_PR_2017} process all proposals together where every proposal shares same CNN layers to allow object classification and discovery at the same time. The \emph{second} kind does not depend on an external instance generator, rather produces the bag using the network itself based on activations of different layers \cite{Feng_AAAI_2017} or from different frames in case of videos \cite{Zhou_2017_ICCV}. These approaches do not have the advantage to consider localized image feature. In our work, we propose a model that combines both kind of approaches to leverage their individual strengths. Our proposed model can generate the bag of instance-patches/proposals by itself without the need of external bag generator and can process all the patches together.

\subsubsection{Zero-shot image tagging:} Instead of assigning one unseen label to an image during recognition task, zero-shot tagging allows to tag multiple unseen tags to an image and/or ranking the array of unseen tags. Although interesting, this problem is not well-addressed in the zero-shot learning literature. An early attempt of such kind extended a work for zero-shot recognition \cite{norouzi_arXiv_2013} to perform zero-shot tagging by proposing a hierarchical semantic embedding to make the label embedding more reliable \cite{Li_tagging_2015}. \cite{Fu_Transductive_2015} proposed a transductive multi-label version of the problem where a predefined and relatively small set of unseen tags were considered. In a recent work, \cite{Zhang_2016_CVPR} proposed a fast zero-shot tagging approach that can be trained using only seen tags and tested using both seen and unseen tags for test images. \cite{Ren_BMVC_17} proposed a multi-instance visual-semantic embedding approach that can extract localized image features. The main drawback of these early efforts is the dependence on pre-trained CNN features (in \cite{Fu_Transductive_2015,Zhang_2016_CVPR,Li_tagging_2015}) or fast-RCNN \cite{FastRCNN_2015_ICCV} features (in \cite{Ren_BMVC_17}) and therefore not end-to-end trainable. In this work, we propose a fully end-to-end solution for both conventional and zero-shot tagging.

\section{Our Method}

\subsection{Problem Formulation}\label{sec:Problem Formulation}

Suppose, we have a set of `seen' tags denoted by $\mathcal{S} = \{1,\ldots, \mathrm{S}\}$ and another set of `unseen' tags $\mathcal{U} = \{\mathrm{S}+1, \ldots,\mathrm{S}+\mathrm{U}\}$, such that  $\mathcal{S} \cap \mathcal{U} = \phi$ where, $\mathrm{S}$ and $\mathrm{U}$ represents the total number of seen and unseen tags respectively. We also denote 
$\mathcal{C} = \mathcal{S} \cup \mathcal{U} $, such that $\mathrm{C} = \mathrm{S} + \mathrm{U}$ is the cardinality of the tag-label space. For each of the tag $c \in \mathcal{C}$, we can obtain a `$d$' dimensional word vector $\mathbf{v}_c$ (word2vec or GloVe) as semantic embedding. The training examples can be defined as a set of tuples, $\{(\mathbf{X}_s, \mathbf{y}_s): s  \in [1,M]\}$, where $\mathbf{X}_s$ is the $s^{th}$ input image and $\mathbf{y}_s \subset \mathcal{S}$ is the set of relevant seen tags.  We represent $u^{th}$ testing image as $\mathbf{X}_u$ which correspondences to a set of relevant seen and/or unseen tag $\mathbf{y}_u \subset \mathcal{C}$. Note that, $\mathbf{X}_u$, $\mathbf{y}_u$, $\mathcal{U}$ and its corresponding word vectors are not observed during training. Now, we define the following problems:

\renewcommand{\labelitemi}{$\bullet$}
\begin{itemize}
\item {\em Conventional tagging:} Given $\mathbf{X}_u$ as input, assign relevant seen tags $\mathbf{y} = \{\mathbf{y}_u \cap \mathcal{S}\}$.

\item {\em Zero-shot tagging (ZST):} Given $\mathbf{X}_u$ as input, assign relevant unseen tags $\mathbf{y} = \{\mathbf{y}_u \cap \mathcal{U}\}$.

\item {\em Generalized zero-shot tagging (GZST):} Given $\mathbf{X}_u$ as input, assign relevant tags from both seen and unseen $\mathbf{y}_u \subset \mathcal{C}$.
\end{itemize}

\noindent \textbf{MIL formulation:} We formulate the above mentioned problems in Multiple Instance Learning (MIL) framework. Let us represent the $s^{th}$ training image with a bag of $n+1$ instances $\mathbf{X}_s = \{ \mathbf{x}_{s0} \ldots \mathbf{x}_{sn}\}$, where $i^{th}$ instance $\mathbf{x}_{si}$ represents either an image patch (for $i > 0$) or the complete image itself (for $i=0$). As $\mathbf{y}_s$ represents relevant seen tags of $\mathbf{X}_s$, according to MIL assumption, the bag has at least one instance for each tag in the set $\mathbf{y}_s$ and no instance for $\mathcal{S} \setminus \mathbf{y}_s$ tags. Thus, instances in $\mathbf{X}_s$ can work as a positive example for $y \in \mathbf{y}_s$ and negative example for $y' \in \{\mathcal{S} \setminus \mathbf{y}_s\}$. This formulation does not use instance level tag annotation which makes it a weakly supervised problem. Our aim is to design and learn an end-to-end deep learning model that can itself generate the appropriate bag of instances and simultaneously assign relevant tags to the bag.

\begin{figure}[t] 
  \centering
   \includegraphics[width=1\textwidth,trim={3.5cm .5cm 1.5cm 0cm},clip]{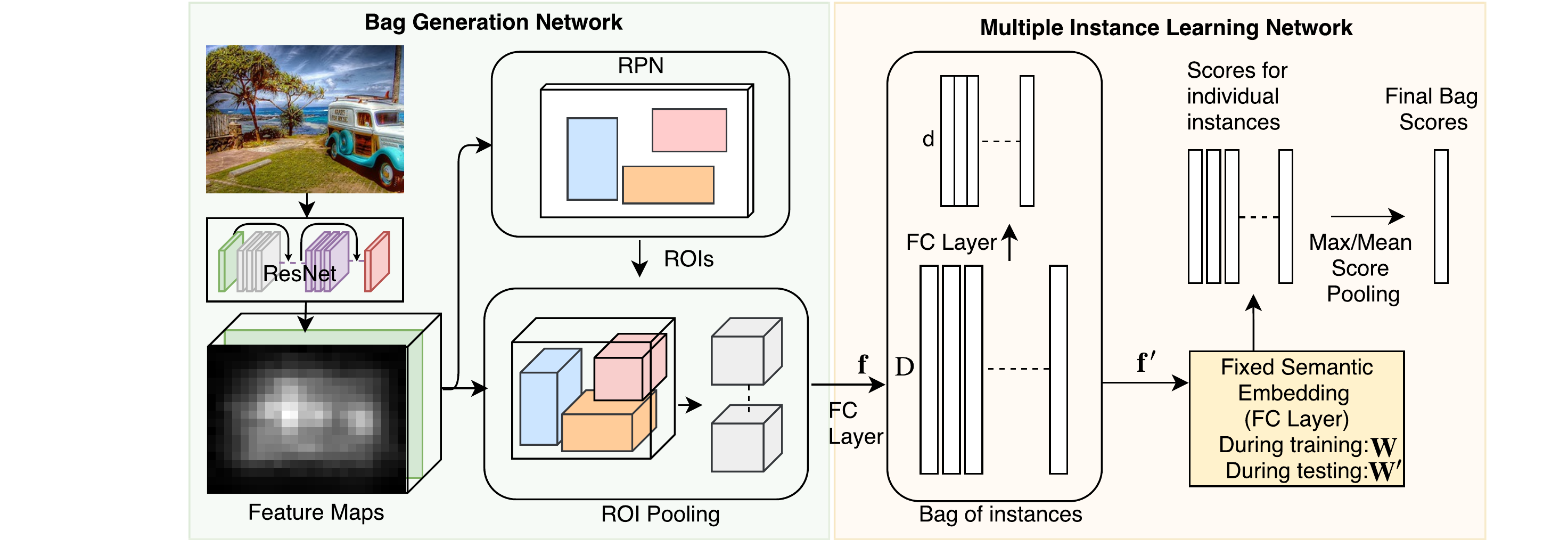}
   \vspace{-1.5em}
   \caption{Proposed network architecture for MIL based zero-shot image tagging.}
\label{fig:network}
\end{figure}

\subsection{Network Architecture}

The proposed network architecture is illustrated in Fig. \ref{fig:network}. It is composed of two parts: bag generation ({\em left}) and multiple instance learning (MIL) network ({\em right}). The bag generation network generates the bag of instances as well as their visual features, and the MIL network processes the resulting bag of instance features to find the final multi-label prediction which is calculated by a global pooling of the prediction scores of individual instances. In this manner, bag generation and zero-shot prediction steps are combined in a unified framework that effectively transfers learning from seen to unseen tags. 

\subsubsection{Bag generation:} In our proposed method, the bag contains image patches which are assumed to cover all objects and concepts presented inside the image. Many closely related traditional methods \cite{Wu_CVPR_2015,Wei_PAMI_2016,TANG_PR_2017,Ren_BMVC_17} apply some external procedure like Selective Search \cite{SS_2013}, EdgeBoxes \cite{Edge_Boxes_2014}, BING \cite{cheng_2014_bing} etc. for this purpose. Such a strategy creates three problems: \textbf{(1)} the off-line external process does not allow an end-to-end learnable framework, \textbf{(2)} patch generation process is prone to more frequent errors because it can not get fine-tuned on the target dataset, and \textbf{(3)} the MIL framework need to process patches rather than the image itself. In this paper, we propose to solve these problems by generating useful bag of patches by the network itself. The recent achievements of object detection framework, such as the Faster-RCNN \cite{Faster_RCNN_2017}, allows us to generate object proposals and later perform detection within a single network. We adopt this strategy to generate bag of image patches for MIL. Remarkably, the original Faster-RCNN model is designed for supervised learning, while our MIL framework extends it to weakly supervised setting. 

A Faster-RCNN model \cite{Faster_RCNN_2017} with Region Proposal Network (RPN) is learned using the ILSVRC-2017 detection dataset. This architecture uses a base network ResNet-50  \cite{ResNet_CVPR_2016} which is shared between RPN and classification/localization network. As practiced, the base network is initialized with the pre-trained weights. Though not investigated in this paper, other popular CNN models e.g., VGG \cite{Vgg_arXiv_2014} and GoogLeNet \cite{GNet_CVPR_2015} can also be used as the shared base network. Now, given a training image $\mathbf{X}_s$, the RPN can produce a fixed number ($n$) of region of interest (ROI) proposals $\{ \mathbf{x}_{s1} \ldots \mathbf{x}_{sn}\}$ with a high recall rate. For image tagging, all tags may not represent an object. Rather, tags can be concepts that describe the whole image e.g.,  nature and landscape. To address this issue, we add a global image ROI (denoted by $\mathbf{x}_{s0}$) comprising of complete image to the ROI proposal set generated by the RPN. Afterwards, ROIs are fed to  ROI-Pooling and subsequent densely connected layers to calculate a $D$-dimensional feature $\mathcal{F}_s = [ \mathbf{f}_{s0} \ldots \mathbf{f}_{sn}] \in \mathbb{R}^{D \times (n+1)}$ for each ROI where $\mathbf{f}_{s0}$ is the feature representation of the whole image. This bag is forwarded to MIL network for prediction.

\subsubsection{Training:} We first initialize a Faster-RCNN model with the pre-trained weights for ILSVRC-2017 object detection dataset. After that, the last two layers (i.e. the classification and localization head) are popped out to produce bag generation network. Our network design then comprises of three fully connected (FC) layers and a global pooling layer (either max or mean pooling) with the resulting network to predict scores for $\mathrm{S}$ number of seen tags. The network is then fine-tuned on target tagging dataset i.e. NUS-WIDE \cite{NUS_WIDE_09}. Note that, at this stage, the resulting network cannot train the RPN further because no localization branch is left in the network. However, the shared base network (ResNet-50) gets fine-tuned for better proposal generation. Similarly, the weights connecting the newly added layers remains trainable along with the other parts of the network. The only exception is the last FC layer, where we use a \textbf{fixed} semantic embedding $\mathbf{W} = [\mathbf{v}_1 \ldots \mathbf{v}_\mathrm{S}] \in \mathbb{R}^{d \times \mathrm{S}}$ containing word vector of seen tags as non-trainable weights for that layer. Notably, a non-linear activation ReLU is applied between the first two FC layer but not considered between second and third layers. The role of the first two layers is to process bag of features by calculating $\mathcal{F'}_s = [ \mathbf{f'}_{s0} \ldots \mathbf{f'}_{sn}] \in \mathbb{R}^{d \times (n+1)}$ for the projection onto the fixed embedding. The third FC layer projects $\mathcal{F'}_s$ in the semantic tag space and produce scores for individual instances. Given, a bag of instance feature $\mathcal{F'}_s$, we compute the prediction scores of individual instances, $\mathbf{P}_s = [ \mathbf{p}_{s0} \ldots \mathbf{p}_{sn}] \in \mathbb{R}^{\mathrm{S} \times (n+1)}$ as follows:

    \begin{equation}
		\mathbf{P}_s = \mathbf{W}^T \mathcal{F'}.
        \label{eq:projection}
    \end{equation}
Often bag generation network produces noisy ROIs which usually results in unreliable scores in $\mathbf{P}_s$. To suppress this noise, we add the global pooling (max or mean) layer which performs the following operation to calculate final multi-label prediction score for a bag:

    \begin{equation}
		\mathbf{o}_s = \max \Big( \mathbf{p}_{s0},\mathbf{p}_{s1}, \ldots , \mathbf{p}_{sn} \Big) \qquad or \qquad \mathbf{o}_s = \frac{1}{n+1}\sum_{j = 0}^{n} \mathbf{p}_{sj}.
    \end{equation}

\subsubsection{Loss formulation:} Suppose, for $s^{th}$ training image, $\mathbf{o}_s = [o_1 \ldots o_{\mathrm{S}} ]$ contains final multi-label prediction of a bag for seen classes. This bag is a positive example for each tag $y \in \mathbf{y}_s$ and negative examples for each tag $y' \in \{\mathcal{S} \setminus \mathbf{y}_s\}$. Thus, for each pair of $y$ and $y'$, the difference $o_{y'} - o_y$ represents the disparity between predictions for positive and negative tags. Our goal is to minimize these differences in each iteration. We formalize the loss of a bag considering it to contain both positive and negative examples for different tags:
\begin{align} 
L_{tag}(\mathbf{o}_s,\mathbf{y}_s) &= \frac{1}{|\mathbf{y}_{s}| |\mathcal{S} \setminus \mathbf{y}_s|} \sum_{y' \in \{\mathcal{S} \setminus \mathbf{y}_s\}} \sum_{y \in \mathbf{y}_s} \log\Big(1 + \exp   (o_{y'} - o_y) \Big). \notag
\end{align}
We minimize the overall loss on all training images as follows:
\begin{equation*}
L = \underset{\Theta}{\arg\min} \frac{1}{M} \sum_{s=1}^M \Big( L_{tag}(\mathbf{o}_s,y_s)\Big).
\end{equation*}
Here, $\Theta$ denote the parameter set of the proposed network and $M$ is the total number of training images.

\subsubsection{Prediction:} During testing, we modify the fixed embedding $\mathbf{W}$ with both seen and unseen word vectors instead of only seen word vectors. Suppose, after modification $\mathbf{W}$ becomes $\mathbf{W}' = [\mathbf{v}_1 \ldots \mathbf{v}_\mathrm{S},  \mathbf{v}_\mathrm{S+1} \ldots \mathbf{v}_{\mathrm{S}+\mathrm{U}}] \in \mathbb{R}^{d \times \mathrm{C}}$. With the use of $\mathbf{W}'$ in Eq. \ref{eq:projection}, we get prediction score of both seen and unseen tags for each individual instance in the bag. Then, after the global pooling (mean or max), we get the final prediction score for each seen and unseen tags. Finally, based on the tagging task (conventional/zero-shot/generalized zero-shot), we assign top $K$ target tags (from the set $\mathcal{S}$, $\mathcal{U}$ or $\mathcal{C}$) with higher scores.


\section{Experiment}

\subsection{Setup}

\subsubsection{Dataset:} We perform our experiments using a real-world web image dataset namely NUS-WIDE \cite{NUS_WIDE_09}. It contains 269,648 images with three sets of per image tags from Flickr. The first, second and third set contain 81, 1000 and 5018 tags respectively. The tags inside the first set are carefully chosen, therefore less noisy whereas third set has the highest noise in annotations. Following the previous work \cite{Zhang_2016_CVPR}, we use 81 first set tags as unseen in this paper. We notice that the tag `interesting' comes twice within the second set. After removing this inconsistency and selecting 81 unseen tags from the second set results in 924 tags which we use as seen for our experiment. The dataset provides the split of 161,789 training and 107,859 testing images. We use this recommended setting ignoring the non-tagged images.
    
\subsubsection{Visual and semantic embedding:} Unlike previous attempts in zero-shot tagging \cite{Li_tagging_2015,Zhang_2016_CVPR},  our model works in an end-to-end manner using ResNet-50 \cite{ResNet_CVPR_2016} as a base network. It means the visual feature are originating from ResNet-50, but they are updated during iterations. As the semantic embedding, we use $\ell_2$ normalized 300 dimensional GloVe vectors \cite{Jeffrey_Glove_2014}. We are unable to use word2vec vectors \cite{Mikolov_NIPS_2013} because the pre-trained word-vector model cannot provide vectors for all of 1005 (924 seen + 81 unseen) tags.

\subsubsection{Evaluation metric:} Following the work \cite{Zhang_2016_CVPR}, we calculate precision (P), recall (R) and F-1 score (F1) of the top $K$ predicted tags ($K=3$ and $5$ is used) and Mean image Average Precision (MiAP) as evaluation metric. We use the following equation to calculate MiAP of an input image $I$:
$$
MiAP(I) = \frac{1}{|\mathcal{G}|} \sum_{j=1}^{|\mathcal{G}|} \frac{r_j}{j} \delta(I,t_j),
$$ 
where, $|\mathcal{G}|=$ total number of ground truth tags, $r_j =$ number of relevant tags of $j^{th}$ rank and $\delta(I,t_j) = 1$ if $j^{th}$ tag $t_j$ is associated with the input image $I$, otherwise $\delta(I,t_j) = 0$.

\subsubsection{Implementation details:} We used the following settings during Faster-RCNN  training, following the proposed settings of \cite{Faster_RCNN_2017}: rescaling shorter size of image as 600 pixels, RPN stride = 16, three anchor box scale 128, 256 and 512 pixels, three aspect ratios 1:1, 1:2 and 2:1, non-maximum suppression (NMS) with IoU threshold = 0.7 with maximum proposal = 300 for faster-RCNN. During training our MIL framework, we generated one bag of instances at each iteration from an image to feed our network. We chose a fixed $n$ number of RoIs proposed by RPN which archives best objectness score. We carried out 774k training iterations using Adam optimizer with a learning rate of $10^{-5}$, $\beta_1 = 0.9$ and $\beta_2 = 0.999$. We implemented our model in \textit{Keras} library.
    
\subsection{Tagging Performance}
In this subsection, we evaluate the performance of our framework on three variants of tagging as introduced in Sec.~\ref{sec:Problem Formulation}, namely conventional, zero-shot and generalized zero-shot tagging.

\subsubsection{Compared methods:} To compare our results, we have reimplemented two similar published methods (ConSE \cite{norouzi_arXiv_2013} and Fast0Tag \cite{Zhang_2016_CVPR}) and one simple baseline based on ResNet-50. We choose these methods for comparison because of their suitability to perform zero-shot tasks. 

ConSE \cite{norouzi_arXiv_2013} was originally introduced for zero-shot learning. This approach first learns a classifier for seen tags and generates a semantic embedding for unseen input by linearly combining word vectors of seen classes using seen prediction scores. In this way, it can rank unseen tags based on the distance of generated semantic embedding and the embedding of unseen tags. 

Fast0Tag \cite{Zhang_2016_CVPR} is the main competitor of our work. This is a deep feature based approach, where features are calculated from a pre-trained VGG-19 \cite{Vgg_arXiv_2014}. Afterwards, a neural network is trained on these features to classify seen and unseen input. This approach outperforms many established methods like WRAP \cite{Gong_ICLR_2013}, WSABIE \cite{WSABIE_IJCAI_2011}, TagProp \cite{TagProp_ICCV_2009}, FastTag \cite{FastTag_ICML_2013} on conventional tagging task. Therefore, in this paper, we do not consider those low-performing methods for comparison. The performance reported in this paper using Fast0Tag method is relatively different from the published results because of few reasons: (1) We use the recent ResNet-50 whereas \cite{Zhang_2016_CVPR} reported results on VGG-19, (2) Although \cite{Zhang_2016_CVPR} experimented on NUS-WIDE, they only used a subset 223,821 images in total, (3) The implementation for \cite{Zhang_2016_CVPR} did not consider the repetition of the seen tag `interesting'.

The baseline method is a special case of our proposed method which uses the whole image as a single instance inside the bag. It breaks the multiple instance learning consideration but does not affect the end-to-end nature of the solution.

\begin{table}[!t]
  \begin{center}
  \scalebox{1}{
    \begin{tabular}{|c|c|c|c|c|c|c|c|}
    \hline
    \multirow{2}{*}{Method}&\multirow{2}{*}{MiAP}&\multicolumn{3}{|c|}{$K=3$}&\multicolumn{3}{|c|}{$K=5$} \\ \cline{3-8}
    &&P&R&F1&P&R&F1\\ \hline
    Fast0Tag \cite{Zhang_2016_CVPR} &35.73&20.24&34.48&25.51&16.16&45.87&23.90 \\ \hline
    Baseline &40.45&22.95&39.09&28.92&17.99&51.09&26.61 \\ \hline
    Ours (Bag: 32)&\textbf{53.97}&30.17&51.41&38.03&22.66&64.35&33.52 \\ \hline
    Ours (Bag: 64)&53.94&\textbf{30.23}&\textbf{51.61}&\textbf{38.18}&\textbf{22.73}&\textbf{64.54}&\textbf{33.62} \\ \hline
    \end{tabular}}
  \end{center}\vspace{-0.5em}
  \caption{Results for conventional tagging. $K$ denotes the number of assigned tags.}
  \label{tab:conventionresult}
\end{table}

\begin{table}[t]
  \begin{center}
  \scalebox{.78}{
    \begin{tabular}{|c|c|c|c|c|c|c|c|c|c|c|c|c|c|c|}
    \hline
    \multirow{3}{*}{Method}&\multicolumn{7}{|c|}{Zero-shot tagging}&\multicolumn{7}{|c|}{Generalized zero-shot tagging} \\ \cline{2-15}
    &\multirow{2}{*}{MiAP}&\multicolumn{3}{|c|}{K=3}&\multicolumn{3}{|c|}{K=5}&\multirow{2}{*}{MiAP}&\multicolumn{3}{|c|}{K=3}&\multicolumn{3}{|c|}{K=5} \\ \cline{3-8} \cline{10-15}
    &&P&R&F1&P&R&F1&&P&R&F1&P&R&F1 \\ \hline
     ConSE \cite{norouzi_arXiv_2013}   &18.91&8.39&14.30&10.58& 7.16&20.33&10.59&7.27&2.11&3.59&2.65& 8.82&5.69&6.92 \\ \hline
    Fast0Tag \cite{Zhang_2016_CVPR}&24.73&13.21&22.51&16.65&11.00&31.23&16.27&10.36&5.21&8.88&6.57&12.41&8.00&9.73 \\ \hline
    Baseline&29.75&16.64&28.34&20.97&13.49&38.32&19.96&12.07&5.99&10.20&7.54&14.28&9.21&11.20 \\ \hline 
    Ours (Bag: 32) &37.50	&21.16	&36.06	&26.67	&16.62	&47.20	&24.59 & \textbf{20.55} & \textbf{27.66}  & \textbf{10.70} & \textbf{15.43} & \textbf{23.67} & \textbf{15.27} & \textbf{18.56} \\ \hline
    Ours (Bag: 64) &\textbf{39.01}	& \textbf{22.05}	& \textbf{37.56}	& \textbf{27.79}	& \textbf{17.26}	& \textbf{49.01}	& \textbf{25.53} & 20.32 &27.09 &10.48&15.12&23.27&15.01&18.25\\ \hline    
    \end{tabular}}
  \end{center}\vspace{-0.5em}
  \caption{Results for zero-shot and generalize zero-shot tagging tasks. }
  \label{tab:zstresult}
\end{table}

\subsubsection{Results:} As mentioned earlier, we perform our training with 924 seen tags and testing with 81 unseen tags for zero-shot settings. However, in conventional tagging case, all tags are considered as seen. Therefore, we use the 81 tag set in both training and testing. Note that, in all of our experiments the same test images are used. Thus, the basic difference between conventional vs. zero-shot tagging is whether those 81 tags were used during training or not. For generalized zero-shot tagging case same testing image set is used, but instead of predicting tags from 81 tag set, our method predicts tags from seen 924 + unseen 81 = 1005 tag set. The performances of ours and compared methods on the tagging tasks are reported in Table \ref{tab:conventionresult} and \ref{tab:zstresult} for the case of NUS-WIDE dataset. Our method outperforms all competitor methods by a significant margin. Notably, the following observations can be developed from the results: \textbf{(1)} The performance of conventional tagging is much better than zero-shot cases because unseen tags and associated images are not present during training for zero-shot tasks. One can consider that the performance of conventional case is an upper-bound for zero-shot tagging case. \textbf{(2)} Similar to previous work \cite{Xian_CVPR_2017}, the performance for the generalized zero-shot tagging task is even poorer than the zero-shot tagging task. This can be explained by the fact that the network gets biased towards the seen tags and scores low on unseen categories. This subsequently leads to a decrease in performance. \textbf{(3)} Similar to the observation from \cite{Zhang_2016_CVPR}, Fast0Tag beats ConSE \cite{norouzi_arXiv_2013} in zero-shot cases. The main reason is that the ConSE \cite{norouzi_arXiv_2013} does not use semantic word vectors during its training which is crucial to find a bridge between seen and unseen tags. No results are reported with ConSE for the conventional tagging case because it is only designed for zero-shot scenarios. \textbf{(4)} The baseline beats other two compared methods across all tasks because of the end-to-end training considering word vectors in the learning phase. This approach is benefited by the appropriate adaptation of feature representations for the tagging task. \textbf{(5)} Our approach outperforms all other competitors because it utilizes localized image features based on MIL, perform end-to-end training and integrate semantic vectors of seen tags within the network. We also illustrate some qualitative comparisons in Fig.~\ref{fig:example}.

\subsubsection{Ablation study:} The proposed framework can work for different number of instances in the bag. Moreover, the global pooling before the last layer can be based on either mean or max pooling. In Table \ref{tab:bagsize}, we perform an ablation study for zero-shot tagging based on different combinations of network settings. We observe that with only a few number of instances (e.g., 4) in the bag, our method can beat state-of-the-art approaches \cite{Zhang_2016_CVPR,norouzi_arXiv_2013}. The required bag size is actually depended on the dataset and pooling type. We notice that a large bag-size improves tagging performance for mean pooling and vise versa for max pooling case. This variation is related to the noise inside the tag annotation of the ground truth. Many previous deep MIL networks \cite{Wei_PAMI_2016,TANG_PR_2017} recommended max-pooling for MIL where they experimented on object detection dataset containing the ground-truth annotation without any label noise. In contrast, other than the 81-tag set, NUS-WIDE contains significant noise in the tag annotations. Therefore, mean-pooling with large bag size achieves a balance in the noisy tags, outperforming max-pooling in general for NUS-WIDE. Notably, the bag size of our framework is far less compared to other MIL approaches \cite{Wei_PAMI_2016,TANG_PR_2017}. Being dependent on external bag generator \cite{SS_2013,Edge_Boxes_2014,cheng_2014_bing} previous methods lose control inside the generation process. Thus, a large bag size helps them to get enough proposals to choose the best score in last max-pooling layer. Conversely, our method controls the bag generation network by fine-tuning shared ResNet-50 layers which eventually can relax the requirement of large bag sizes. 

\begin{table}[!t]
  \begin{center}
  \scalebox{.8}{
    \begin{tabular}{|c|c|c|c|c|c|c|c|c|c|c|c|c|c|c|}
    \hline
    \multirow{3}{*}{\makecell{Bag size\\$(n+1)$}}&\multicolumn{7}{|c|}{Mean Pooling}&\multicolumn{7}{|c|}{Max Pooling} \\ \cline{2-15}
    &\multirow{2}{*}{MiAP}&\multicolumn{3}{|c|}{K=3}&\multicolumn{3}{|c|}{K=5}&\multirow{2}{*}{MiAP}&\multicolumn{3}{|c|}{K=3}&\multicolumn{3}{|c|}{K=5} \\ \cline{3-8} \cline{10-15}
    &&P&R&F1&P&R&F1&&P&R&F1&P&R&F1 \\ \hline
    4	&32.85	&18.28	&31.15	&23.04	&14.79	&42.00	&21.88 	& \textbf{35.75}	& \textbf{20.20}	& \textbf{34.42}	& \textbf{25.46}	& \textbf{16.13}	& \textbf{45.79}	& \textbf{23.85} \\ \hline
    8	&36.33	&20.68	&35.23	&26.06	&16.38	&46.51	&24.23  &34.89	&19.61	&33.42	&24.72	&15.67	&44.49	&23.18\\ \hline
    16  &37.20	&21.08	&35.92	&26.57	&16.58	&47.08	&24.53  &32.03	&18.01	&30.68	&22.69	&14.85	&42.17	&21.97\\ \hline 
    32	&37.50	&21.16	&36.06	&26.67	&16.62	&47.20	&24.59  &29.29	&16.57	&28.23 	&20.88	&14.00	&39.75	&20.70\\ \hline
    64	& \textbf{39.01}	& \textbf{22.05}	& \textbf{37.56}	& \textbf{27.79}	& \textbf{17.26}	& \textbf{49.01}	& \textbf{25.53}  &32.05	&17.68	&30.13	&22.29	&14.38	&40.82	&21.27 \\ \hline
    \end{tabular}}
  \end{center}\vspace{-0.5em}
  \caption{Ablation study: Impact of pooling type and bag size on zero-shot tagging}
  \label{tab:bagsize}
\end{table}

\subsubsection{Tagging in the wild:} Since our method does not use any information about unseen tags in zero-shot settings, it can process an infinite number of unseen tags from an open vocabulary. We test such setting using 5018 tag set of NUS-WIDE. We remove 924 seen tags and ten other tags (handsewn, interestingness, manganite, marruecos, mixs, monochromia, shopwindow, skys, topv and uncropped for which no GloVe vectors were found) to produce a large set of 4084 unseen tags. {After training with 924 seen tags}, the performance of zero-shot tagging with this set is shown in Table \ref{tab:zst5000}. Because of extreme noise in the annotations, the results are very poor in general, but our method still outperforms other competitors \cite{Zhang_2016_CVPR,norouzi_arXiv_2013}.

\begin{table}[!t]
  \begin{center}
  \scalebox{1}{
    \begin{tabular}{|c|c|c|c|c|c|c|c|}
    \hline
    \multirow{2}{*}{Method}&\multirow{2}{*}{MiAP}&\multicolumn{3}{|c|}{K=3}&\multicolumn{3}{|c|}{K=5} \\ \cline{3-8}
    &&P&R&F1&P&R&F1\\ \hline
	ConSE \cite{norouzi_arXiv_2013}&0.36&0.08&0.06&0.07&0.10&0.13&0.11\\ \hline 
    Fast0Tag \cite{Zhang_2016_CVPR}&3.26&3.15&2.40&2.72&2.51&3.18&2.81 \\ \hline
    Baseline&3.61&3.51&2.67&3.04&2.83&3.59&3.16 \\ \hline
    Ours (Bag: 32) & \textbf{5.85} & \textbf{5.42} & \textbf{4.12} & \textbf{4.68} & \textbf{4.42} & \textbf{5.60} & \textbf{4.94} \\ \hline
    Ours (Bag: 64) & 5.52 & 5.01 & 3.81 & 4.33 & 4.10 & 5.20 & 4.59 \\ \hline
    \end{tabular}}
  \end{center}\vspace{-0.5em}
  \caption{Results for zero-shot tagging task with 4,084 unseen tags.}
  \label{tab:zst5000}
\end{table}

\subsection{Zero Shot Recognition (ZSR)}
Our proposed framework is designed to handle zero-shot multi-label problem. Therefore, it can also be used for single label ZSR problem. To evaluate the performance of ZSR, we experiment with the Caltech-UCSD Birds-200-2011 (CUB) dataset \cite{CUB_2011}. Although the size of this dataset is relatively small containing 11,788 images belonging to 200 classes, it is popular for fine-grain recognition tasks. In ZSR literature \cite{Xian_CVPR_2017,Xian_2016_CVPR}, the standard train/test split allows fixed 150 seen and 50 unseen classes for experiments. We follow this traditional setting without using bounding boxes annotation, per image part annotation (like \cite{Akata_2016_CVPR}) and descriptions (like \cite{Zhang_2017_CVPR}). To be consistent with the rest of the paper, we consider 400-d unsupervised GloVe (glo) and word2vec (w2v) vectors used in \cite{Xian_2016_CVPR}. For a given test image, our network predicts unseen class scores and an image is classified to the unseen class which gets the maximum score. As per standard practice, we report the mean per class Top1 accuracy of unseen classes in Table \ref{tab:cub_shorter}. Our method achieves superior results in comparison to state-of-the-art methods using the same settings. Note that, all other methods are deep feature (VGG/GoogLeNet) based approaches that do not train a joint framework in an end-to-end manner. In contrast, our method is end-to-end learnable based on ResNet-50 and additionally generates bounding boxes without using any box annotations.

\begin{SCtable}[][!t]
  \scalebox{0.85}{
    \begin{tabular}{|c|c|c|c|}
    \hline
    Top1 Accuracy & Network & w2v  & glo \\
    \hline
    Akata'16 \cite{Akata_2016_CVPR} & V &33.90&- \\
    DMaP-I'17\cite{Li_2017_CVPR} & G+V &26.38&30.34 \\ 
    SCoRe'17\cite{Morgado_2017_CVPR} & G &31.51&- \\
    Akata'15 \cite{Akata_CVPR_2015} & G &28.40&24.20 \\
    LATEM'16 \cite{Xian_2016_CVPR} & G &31.80&32.50 \\
    DMaP-I'17 \cite{Li_2017_CVPR} & G &26.28&23.69 \\
    \hline    
    Ours (Bag size: 32) & R & 31.77 & 29.56 \\
	Ours (Bag size: 64) & R & \textbf{36.55} & \textbf{33.00} \\ 
    \hline
    \end{tabular}}
  \caption{Zero shot recognition on CUB using mean pooling based MIL. For fairness, we only compared our results with the inductive setting of other methods without per image part annotation and description. We refer investigated network structures as V=VGG, R=ResNet, G=GoogLeNet. 
}
  \label{tab:cub_shorter}
\end{SCtable}



\begin{figure}[t]
  \centering
   \includegraphics[width=1.0\textwidth]{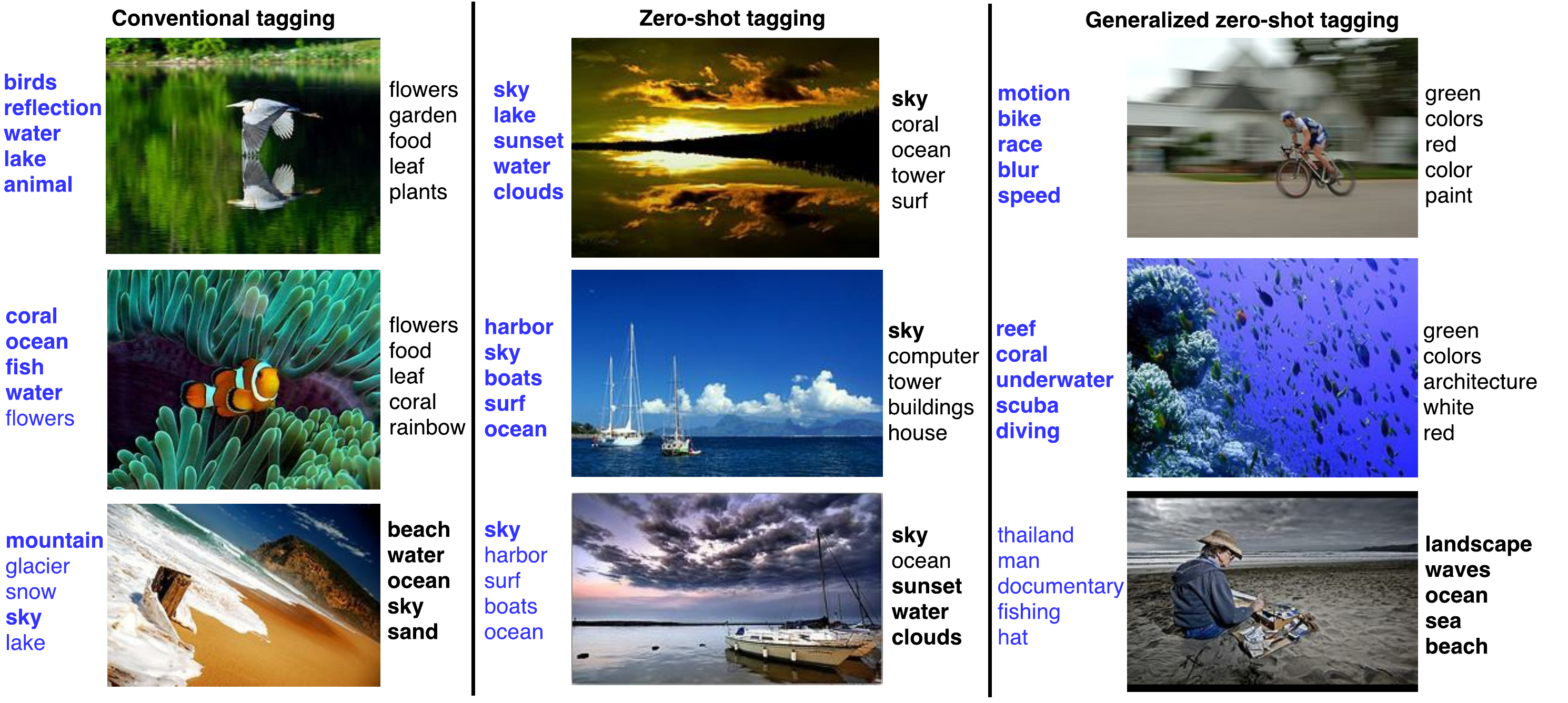}
   \vspace{-2em}
   \caption{Example of top 5 predicted tags across different tasks by our method (left/blue) and fast0tag \cite{Zhang_2016_CVPR} (right/black). \textbf{Bold} text represents the correct tags according to ground-truth. First two rows of images illustrate successful examples of our method and third row is for negative cases.}
\label{fig:example}
\end{figure}

\begin{figure}[t]
  \centering
   \includegraphics[width=1\textwidth]{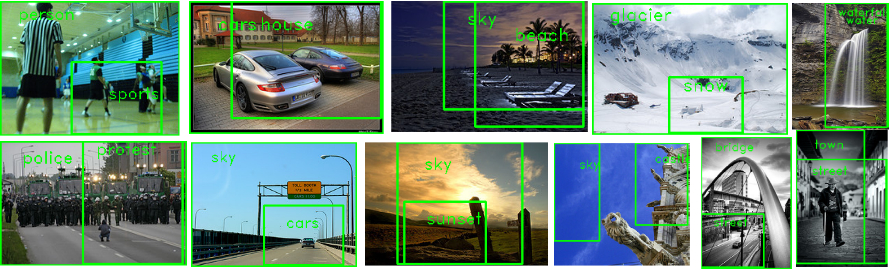}
   \vspace{-2em}
   \caption{Tag discovery. Bounding boxes are shown for Top 2 tags.}
\label{fig:box}
\end{figure}

\subsection{Discussion}

\subsubsection{How does MIL help in the multi-label zero-shot annotation?} We explain this aspect using the illustration in Figure \ref{fig:example}. One can observe that several tags pertain to localized information in a scene that is represented  by only a small subset of the whole image, e.g., fish, coral, bike and bird. This demonstrates that a multi-label tagging method should consider localized regions in conjunction with the whole image. Our proposed method incorporates such consideration using MIL. Therefore, it can annotate those localized tags where previous method, fast0tag, \cite{Zhang_2016_CVPR} usually fails (see rows 1-2 in Fig.~\ref{fig:example}). However, tags like beach, sunset, landscape in the third row of the figure are related to the global information in an image which does not depend on the localized features. Therefore, in this respect, our method sometimes fail in compared to fast0tag \cite{Zhang_2016_CVPR} (see row 3 in Fig.~\ref{fig:example}). However, as illustrated in Fig.~\ref{fig:example} (the non-bold tags in blue and black colors), the predicted tags of our method in those failure cases are still meaningful and relevant compared to the failure cases of fast0tag \cite{Zhang_2016_CVPR}. 

\subsubsection{Image location and tag correspondence:} As a byproduct, our approach can generate a bounding box for each assigned tag. In Fig.~\ref{fig:box}, we illustrate some boxes (for top 2 tags) to indicate the correspondence between image locations and associated tags. Note that, our method often selects the whole image as one bounding box because we consider whole image as an instance inside the bag. This consideration is particularly helpful for NUS-WIDE dataset because it contains many tags which are not only related to objects but are relevant to the overall scene such as natural concept (sky, water, sunset), aesthetic style (reflection, tattoo) or action (protest, earthquake, sports). Any quantitative analysis for this weakly supervised box detection task was not possible because the NUS-WIDE dataset does not provide any localization ground-truth for tags in an image. 


\section{Conclusion}
While traditional zero-shot learning methods only handle single unseen label per image, this paper attempts to assign multiple unseen tags. For the first time, we propose an end-to-end, deep MIL framework to tackle multi-label zero-shot tagging. Unlike previous models for traditional image tagging, our MIL framework does not depend on off-line bag generator. Building on recent advancements in object detection, our model automatically generates the bag of instances in an efficient manner and can assign both seen and unseen labels to input images. Moreover, any number of unseen tags from an open vocabulary could be employed in the model during test time. In addition, our method can be viewed as a weakly supervised learning approach because of its ability to find a bounding box for each tag without requiring any box annotation during training. We validate our framework by achieving state-of-the-art performance on a large-scale tagging dataset outperforming established methods in the literature. In future, we will explore the semantic relationship of word vectors to incorporate dependency among tags.

\clearpage

\bibliographystyle{splncs}
\bibliography{egbib}

\begin{thebibliography}{10}

\bibitem{Xian_CVPR_2017}
Xian, Y., Schiele, B., Akata, Z.:
\newblock Zero-shot learning - the good, the bad and the ugly.
\newblock In: IEEE Computer Vision and Pattern Recognition (CVPR). (2017)

\bibitem{Kodirov_2017_CVPR}
Kodirov, E., Xiang, T., Gong, S.:
\newblock Semantic autoencoder for zero-shot learning.
\newblock In: The IEEE Conference on Computer Vision and Pattern Recognition
  (CVPR). (July 2017)

\bibitem{Zhang_2017_CVPR}
Zhang, L., Xiang, T., Gong, S.:
\newblock Learning a deep embedding model for zero-shot learning.
\newblock In: The IEEE Conference on Computer Vision and Pattern Recognition
  (CVPR). (July 2017)

\bibitem{Deutsch_2017_CVPR}
Deutsch, S., Kolouri, S., Kim, K., Owechko, Y., Soatto, S.:
\newblock Zero shot learning via multi-scale manifold regularization.
\newblock In: The IEEE Conference on Computer Vision and Pattern Recognition
  (CVPR). (July 2017)

\bibitem{Li_2017_CVPR}
Li, Y., Wang, D., Hu, H., Lin, Y., Zhuang, Y.:
\newblock Zero-shot recognition using dual visual-semantic mapping paths.
\newblock In: The IEEE Conference on Computer Vision and Pattern Recognition
  (CVPR). (July 2017)

\bibitem{Morgado_2017_CVPR}
Morgado, P., Vasconcelos, N.:
\newblock Semantically consistent regularization for zero-shot recognition.
\newblock In: The IEEE Conference on Computer Vision and Pattern Recognition
  (CVPR). (July 2017)

\bibitem{Akata_2016_CVPR}
Akata, Z., Malinowski, M., Fritz, M., Schiele, B.:
\newblock Multi-cue zero-shot learning with strong supervision.
\newblock In: The IEEE Conference on Computer Vision and Pattern Recognition
  (CVPR). (June 2016)

\bibitem{Mensink_CVPR_2014}
Mensink, T., Gavves, E., Snoek, C.G.:
\newblock Costa: Co-occurrence statistics for zero-shot classification.
\newblock In: Proceedings of the IEEE Conference on Computer Vision and Pattern
  Recognition. (2014)  2441--2448

\bibitem{Fu_Transductive_2015}
Fu, Y., Yang, Y., Hospedales, T., Xiang, T., Gong, S.:
\newblock Transductive multi-label zero-shot learning.
\newblock arXiv preprint arXiv:1503.07790 (2015)

\bibitem{Zhang_2016_CVPR}
Zhang, Z., Saligrama, V.:
\newblock Zero-shot learning via joint latent similarity embedding.
\newblock In: The IEEE Conference on Computer Vision and Pattern Recognition
  (CVPR). (June 2016)

\bibitem{Li_tagging_2015}
Li, X., Liao, S., Lan, W., Du, X., Yang, G.:
\newblock Zero-shot image tagging by hierarchical semantic embedding.
\newblock In: Proceedings of the 38th International ACM SIGIR Conference on
  Research and Development in Information Retrieval, ACM (2015)  879--882

\bibitem{Feng_AAAI_2017}
Feng, J., Zhou, Z.H.:
\newblock Deep miml network.
\newblock (2017)  1884--1890 cited By 0.

\bibitem{Zhou_2017_ICCV}
Zhou, Y., Sun, X., Liu, D., Zha, Z., Zeng, W.:
\newblock Adaptive pooling in multi-instance learning for web video annotation.
\newblock In: The IEEE International Conference on Computer Vision (ICCV). (Oct
  2017)

\bibitem{Wang_ICCV_2015}
Wang, X., Zhu, Z., Yao, C., Bai, X.:
\newblock Relaxed multiple-instance svm with application to object discovery.
\newblock Volume 2015 International Conference on Computer Vision, ICCV 2015.
  (2015)  1224--1232 cited By 10.

\bibitem{Wu_CVPR_2015}
Wu, J., Yu, Y., Huang, C., Yu, K.:
\newblock Deep multiple instance learning for image classification and
  auto-annotation.
\newblock In: 2015 IEEE Conference on Computer Vision and Pattern Recognition
  (CVPR). (June 2015)  3460--3469

\bibitem{Wei_PAMI_2016}
Wei, Y., Xia, W., Lin, M., Huang, J., Ni, B., Dong, J., Zhao, Y., Yan, S.:
\newblock Hcp: A flexible cnn framework for multi-label image classification.
\newblock IEEE Transactions on Pattern Analysis and Machine Intelligence
  \textbf{38}(9) (2016)  1901--1907 cited By 22.

\bibitem{TANG_PR_2017}
Tang, P., Wang, X., Huang, Z., Bai, X., Liu, W.:
\newblock Deep patch learning for weakly supervised object classification and
  discovery.
\newblock Pattern Recognition \textbf{71} (2017)  446 -- 459

\bibitem{SS_2013}
Uijlings, J.R.R., van~de Sande, K.E.A., Gevers, T., Smeulders, A.W.M.:
\newblock Selective search for object recognition.
\newblock International Journal of Computer Vision \textbf{104}(2) (Sep 2013)
  154--171

\bibitem{Edge_Boxes_2014}
Zitnick, C.L., Doll{\'a}r, P.:
\newblock Edge boxes: Locating object proposals from edges.
\newblock In Fleet, D., Pajdla, T., Schiele, B., Tuytelaars, T., eds.: Computer
  Vision -- ECCV 2014, Cham, Springer International Publishing (2014)  391--405

\bibitem{cheng_2014_bing}
Cheng, M.M., Zhang, Z., Lin, W.Y., Torr, P.:
\newblock Bing: Binarized normed gradients for objectness estimation at 300fps.
\newblock In: Proceedings of the IEEE conference on computer vision and pattern
  recognition. (2014)  3286--3293

\bibitem{Tang_TIP_2017}
Tang, P., Wang, X., Feng, B., Liu, W.:
\newblock Learning multi-instance deep discriminative patterns for image
  classification.
\newblock IEEE Transactions on Image Processing \textbf{26}(7) (2017)
  3385--3396

\bibitem{FastRCNN_2015_ICCV}
Girshick, R.:
\newblock Fast r-cnn.
\newblock In: The IEEE International Conference on Computer Vision (ICCV).
  (December 2015)

\bibitem{Ren_BMVC_17}
Ren, Z., Jin, H., Lin, Z., Fang, C., Yuille, A.:
\newblock Multiple instance visual-semantic embedding.
\newblock In: Proceeding of the British Machine Vision Conference (BMVC).
  (2017)

\bibitem{Demirel_2017_ICCV}
Demirel, B., Gokberk~Cinbis, R., Ikizler-Cinbis, N.:
\newblock Attributes2classname: A discriminative model for attribute-based
  unsupervised zero-shot learning.
\newblock In: The IEEE International Conference on Computer Vision (ICCV). (Oct
  2017)

\bibitem{Faster_RCNN_2017}
Ren, S., He, K., Girshick, R., Sun, J.:
\newblock Faster r-cnn: Towards real-time object detection with region proposal
  networks.
\newblock IEEE Transactions on Pattern Analysis and Machine Intelligence
  \textbf{39}(6) (June 2017)  1137--1149

\bibitem{Dai_RFCN_2016}
Jifeng~Dai, Yi~Li, K.H.J.S.:
\newblock {R-FCN}: Object detection via region-based fully convolutional
  networks.
\newblock arXiv preprint arXiv:1605.06409 (2016)

\bibitem{Liu_SSD_2016}
Liu, W., Anguelov, D., Erhan, D., Szegedy, C., Reed, S., Fu, C.Y., Berg, A.C.
\newblock In: SSD: Single Shot MultiBox Detector. Springer International
  Publishing, Cham (2016)  21--37

\bibitem{Redmon_yolo9000_2016}
Redmon, J., Farhadi, A.:
\newblock Yolo9000: Better, faster, stronger.
\newblock arXiv preprint arXiv:1612.08242 (2016)

\bibitem{norouzi_arXiv_2013}
Norouzi, M., Mikolov, T., Bengio, S., Singer, Y., Shlens, J., Frome, A.,
  Corrado, G.S., Dean, J.:
\newblock Zeroshot learning by convex combination of semantic embeddings.
\newblock In: In Proceedings of ICLR. (2014)

\bibitem{ResNet_CVPR_2016}
He, K., Zhang, X., Ren, S., Sun, J.:
\newblock Deep residual learning for image recognition.
\newblock Volume 2016-January. (2016)  770--778 cited By 107.

\bibitem{Vgg_arXiv_2014}
Simonyan, K., Zisserman, A.:
\newblock Very deep convolutional networks for large-scale image recognition.
\newblock arXiv preprint arXiv:1409.1556 (2014)

\bibitem{GNet_CVPR_2015}
Szegedy, C., Liu, W., Jia, Y., Sermanet, P., Reed, S., Anguelov, D., Erhan, D.,
  Vanhoucke, V., Rabinovich, A.:
\newblock Going deeper with convolutions.
\newblock Proceedings of the IEEE Computer Society Conference on Computer
  Vision and Pattern Recognition \textbf{07-12-June-2015} (2015)  1--9

\bibitem{NUS_WIDE_09}
Chua, T.S., Tang, J., Hong, R., Li, H., Luo, Z., Zheng, Y.T.:
\newblock Nus-wide: A real-world web image database from national university of
  singapore.
\newblock In: Proc. of ACM Conf. on Image and Video Retrieval (CIVR'09),
  Santorini, Greece. (July 8-10, 2009)

\bibitem{Jeffrey_Glove_2014}
Pennington, J., Socher, R., Manning, C.D.:
\newblock Glove: Global vectors for word representation.
\newblock In: Empirical Methods in Natural Language Processing (EMNLP). (2014)
  1532--1543

\bibitem{Mikolov_NIPS_2013}
Mikolov, T., Sutskever, I., Chen, K., Corrado, G.S., Dean, J.:
\newblock Distributed representations of words and phrases and their
  compositionality.
\newblock In Burges, C.J.C., Bottou, L., Welling, M., Ghahramani, Z.,
  Weinberger, K.Q., eds.: Advances in Neural Information Processing Systems 26.
\newblock Curran Associates, Inc. (2013)  3111--3119

\bibitem{Gong_ICLR_2013}
Gong, Y., Jia, Y., Leung, T., Toshev, A., Ioffe, S.:
\newblock Deep convolutional ranking for multilabel image annotation.
\newblock arXiv preprint arXiv:1312.4894 (2013)

\bibitem{WSABIE_IJCAI_2011}
Weston, J., Bengio, S., Usunier, N.:
\newblock Wsabie: Scaling up to large vocabulary image annotation.
\newblock (2011)  2764--2770

\bibitem{TagProp_ICCV_2009}
Guillaumin, M., Mensink, T., Verbeek, J., Schmid, C.:
\newblock Tagprop: Discriminative metric learning in nearest neighbor models
  for image auto-annotation.
\newblock In: 2009 IEEE 12th International Conference on Computer Vision. (Sept
  2009)  309--316

\bibitem{FastTag_ICML_2013}
Chen, M., Zheng, A., Weinberger, K.Q.:
\newblock Fast image tagging.
\newblock In: Proceedings of the 30th International Conference on Machine
  Learning, ICML (January 2013)

\bibitem{CUB_2011}
Wah, C., Branson, S., Welinder, P., Perona, P., Belongie, S.:
\newblock {The Caltech-UCSD Birds-200-2011 Dataset}.
\newblock Technical Report CNS-TR-2011-001, California Institute of Technology
  (2011)

\bibitem{Xian_2016_CVPR}
Xian, Y., Akata, Z., Sharma, G., Nguyen, Q., Hein, M., Schiele, B.:
\newblock Latent embeddings for zero-shot classification.
\newblock In: The IEEE Conference on Computer Vision and Pattern Recognition
  (CVPR). (June 2016)

\bibitem{Akata_CVPR_2015}
Akata, Z., Reed, S., Walter, D., Lee, H., Schiele, B.:
\newblock Evaluation of output embeddings for fine-grained image
  classification.
\newblock In: Proceedings of the IEEE Computer Society Conference on Computer
  Vision and Pattern Recognition. Volume 07-12-June-2015. (2015)  2927--2936

\end{thebibliography}
\end{document}